\pdfoutput=1

\documentclass[11pt]{article}

\usepackage{acl}

\usepackage{times}
\usepackage{latexsym}
\usepackage{graphicx}
\usepackage{subcaption}

\usepackage[T1]{fontenc}

\usepackage[utf8]{inputenc}

\usepackage{microtype}

\title{Task2Dial: A Novel Task and Dataset for Commonsense enhanced Task-based Dialogue Grounded in Documents}

\author{Carl Strathearn \and Dimitra Gkatzia\\
  Edinburgh Napier University \\
  \texttt{\{c.strathearn,d.gkatzia\}@napier.ac.uk}}

\begin{document}
\maketitle
\begin{abstract}
This paper proposes a novel task on commonsense-enhanced task-based dialogue grounded in documents and describes the Task2Dial dataset, a novel dataset of document-grounded task-based dialogues, where an Information Giver (IG) provides instructions (by consulting a document) to an Information Follower (IF), so that the latter can successfully complete the task. In this unique setting, the IF can ask clarification questions which may not be grounded in the underlying document and require commonsense knowledge to be answered. The Task2Dial dataset poses new challenges: (1) its human reference texts show more lexical richness and variation than other document-grounded dialogue datasets; (2) generating from this set requires paraphrasing as instructional responses might have been modified from the underlying document; (3) requires commonsense knowledge, since questions might not necessarily be grounded in the document; (4) generating requires planning based on context, as task steps need to be provided in order. The Task2Dial dataset contains dialogues with an average $18.15$ number of turns and $19.79$ tokens per turn, as compared to $12.94$ and $12$ respectively in existing datasets. As such, learning from this dataset promises more natural, varied and less template-like system utterances. 
\end{abstract}

\section{Introduction}

Goal and task oriented dialogue systems enable users to complete tasks, such as restaurant reservations and travel booking, through conversation \cite{Chen:2017}. Traditionally, goal-oriented dialogue is based on domain-specific database schemas \cite{shah-etal-2018-bootstrapping}, however, encoding all domain information can be prohibitive since most domain knowledge exists in some unstructured format, such as documents \cite{feng-etal-2020-doc2dial}, grounding dialogue in documents is a promising direction for several tasks. Here, we propose a new task for document-grounded dialogue, \texttt{Task2Dial}, which aims at generating instructions grounded in a document so that the receiver of the instructions can complete a task. This task requires following steps in a pre-specified order, invoking every day communication characteristics, such as asking for clarification, questions or advice, which might require the use of commonsense knowledge. The proposed task is different to existing document-grounded tasks such as CoQA \cite{reddy-etal-2019-coqa} in the sense that it goes beyond question answering grounded in a document, as answers might require commonsense knowledge and the underlying information might not be present in the document. At the same time, this challenging task aims to accommodate task-based dialogue, where the information follower has to comprehend (and confirm) all steps for completing the task. 
\begin{figure}
  \includegraphics[width=0.95\linewidth]{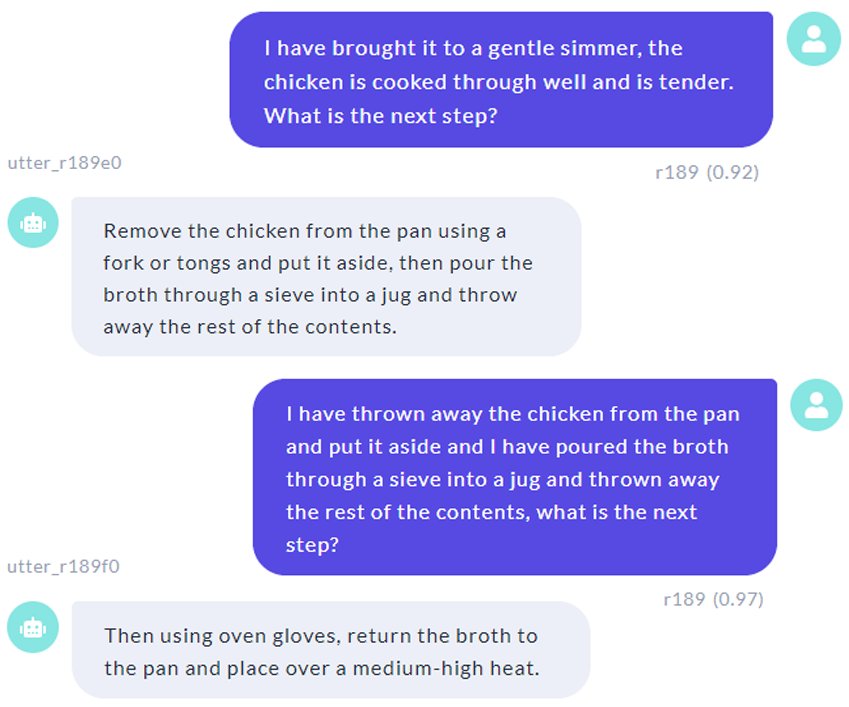}
  \caption{Excerpt from dialogue showing the commonsense handling of hot objects in the Task2Dial dataset.}
  \label{fig:B}
\end{figure}
Inspired by previous work on document-grounded dialogue \cite{feng-etal-2020-doc2dial,hu-etal-2016-corpus,stoyanchev-piwek-2010-constructing}, commonsense-enhanced natural language generation (NLG) \cite{lin-etal-2020-commongen,clinciu-etal-2021-commonsense}, referring expressions generation \cite{PANAGIARIS2021101184}, concept acquisition \cite{Gkatzia-hri-2021}, and task-based/instructional dialogue \cite{gargett-etal-2010-give}, we aim to capture two different types of knowledge: (1) document-level procedural context, i.e. what is the next step; (2) commonsense, i.e. answering questions that are not available in the document, as demonstrated in Figure \ref{fig:B}. The task is designed as an instruction-following scenario with an information giver (IG) and an information follower (IF), inspired partly by the GIVE challenge \cite{gargett-etal-2010-give}. The IG has access to the recipe and gives instructions to the IF. The IG might choose to omit irrelevant information, simplify the wording or provide it as is. The IF will either 'follow' the task by providing confirmation that they have understood the instruction or ask for further information. The IG might have to rely on information outside the given document, in other words the IG will rely on their common sense to enhance understanding and success of the task.
\paragraph{Task Description} The proposed task considers the recipe-following scenario with an information giver (IG) and an information follower (IF), where the IG has access to the recipe and gives instructions to the IF. The IG might choose to omit irrelevant information, simplify the wording in the recipe or provide it as is. The IF will either follow the task or ask for further information. The IG might have to rely on information outside the given document (i.e.\ commonsense) to enhance understanding and success of the task. In addition, the IG decides on how to present the recipe steps, i.e.\ split them into sub-steps or merge them together, often diverting from the original number of recipe steps. The task is regarded successful when the IG has successfully followed/understood the recipe. Hence, other dialogue-focused metrics, such as the number of turns, are not appropriate here. Formally, \textit{Task2Dial} can be defined as follows: Given a recipe $R_i$ from $R = {R_1, R_2, R_3,..., R_n}$, an ontology or ontologies $O_i = {O_11, O_2,..., O_n}$ of cooking related concepts, a history of the conversation $h$, predict the response $r$ of the IG.

This paper follows a theoretical framework which combines a background literature review with the design, development and challenges of the Task2Dial dataset (\S\ref{sec:theoretical-framework}). The proceeding sections cover the data curation methodology (\S\ref{sec:data-curation}), present an analysis of the Task2Dial dataset and a comparison to related datasets (\S\ref{sec:analysis}), discuss the related work (\S\ref{sec:related-work}), and finally discuss the implications and challenges for the development of instruction-giving dialogue systems.  

\section{Theoretical Framework} \label{sec:theoretical-framework}

The proposed task and associated dataset have connections to several lines of research in task and goal oriented dialogue, dialogue tracking and planning, document-grounded dialogue and commonsense reasoning. We next review related work in these areas while grounding our work.

\subsection{Task and Goal-oriented dialogue}

In dialogue management, task-oriented approaches focus on the successful completion of the individual stages of a task, towards achieving an end goal \cite{NEURIPS2020_e9462095}. Comparatively, goal-oriented approaches focus on comparing the outcome or overall performance against a gold standard \cite{ham-etal-2020-end}. Task and goal oriented dialogue systems are common in domains such as booking and reservation systems for businesses \cite{Zhang-recent-2020}. However, business models are typically goal-oriented as the instructions are minimal and the focus is on the outcome \cite{Ilievski:2018}. Instead, the Task2Dial task is formulated as a task-oriented dialogue paradigm to imitate real-world practical scenarios that can vary in complexity and require adaptability, additional information, clarification and natural conversation in order to enhance understanding and success. 

\subsection{Dialogue State Tracking and Planning}

Task-based dialogue systems require the user and artificial agent to work synergistically by following and reciting instructions to achieve a goal. \citet{Zamanirad2020} define these methods in human-bot conversational models as:

\begin{itemize}
    \item \textbf{Single intent and single turn policy:} relies solely on question and answer pairs assuming that the user provides all slot values in a single utterance. This type of task does not require dialogue state tracking.
    \item \textbf{Single intent and multi-turn policy:} Extends the previous conversational model, however this model can include multiple turns, to fill in missing information. Historic information is then extracted from all turns and used to structure data. 
    \item \textbf{Multi-intent and multi-turn policy:} the intents can change  depending on the context.
\end{itemize}

Instruction-giving scenarios follow the \textit{multi-intent multi-turn} conversational framework, since they must accommodate knowledge and variability outside of a linear deterministic model as practical tasks can vary in complexity and the conversation can vary based on the interlocutors prior knowledge. In addition, there is no restriction on the amount of variability introduced into a task, such as introducing alternate methods, commonsense knowledge and concepts that change the structure and information within the dialogue. Variability is often reduced in human-machine scenarios as systems are limited in knowledge and their ability to respond to questions not seen in training \cite{shum2018eliza}, which can result in shortened responses and fewer questions asked on aspects of the task \cite{Byrne2019}. This hinders the system's ability to ensure that the IF has understood the IGs directions, which may produce irregular outcomes or result in an incomplete task. Therefore, capturing and emulating natural variability within the dialogue is crucial for creating robust and reliable conversational systems for instruction-giving scenarios.

Similarly to existing datasets such as Multi-Domain Wizard-of-Oz (MultiWOZ) \cite{budzianowski-etal-2018-multiwoz}, Taskmaster-1 \cite{Byrne2019}, Doc2dial \cite{feng-etal-2020-doc2dial} and the Action-Based Conversations Dataset (ABCD) \cite{chen-etal-2021-action}, Task2Dial also addresses the task of completing a process by following a sequence of steps. However, in addition to grounded information in documents, Task2Dial aims to accommodate questions and clarifications on different aspects of the task that might not be grounded in the document. In previous work, the user is limited to the path of the subroutine, however in Task2Dial, the IF can ask the IG questions at any stage of the task, regardless of the position within a given sequence and then return to that position after the question is fulfilled. For example, in a cooking scenario the IF may ask the IG how to use a certain kitchen utensil. The IG would need to answer this question, then return to the correct stage in the recipe in order to continue the sequence. This introduces additional challenges for state-tracking. The conversational agent must not only generate appropriate sequential instructions based on a document, it must also be able to request confirmation that the user has understood the task, and be able to answer questions outside its pre-defined script. Using document-grounded subroutines to capture intents that change the direction of a task broadens the interaction between the IG and IF \cite{chen-etal-2021-action}, introducing new challenges for dialogue state-tracking. 

\subsection{Document-grounded dialogue}
Document-grounded dialogue systems (DGDS) classify unstructured, semi-structured and structured information in documents to aid understanding human knowledge and interactions, creating greater naturalistic human-computer interactions (HCI) \cite{zhou-etal-2018-dataset}. The aim of DGDS is to formulate a mode of conversation from the information (utterances, turns, context, clarification) provided in a document(s) \cite{Ma-docdial-2020}. DGDS are particularly useful in task-oriented and goal-oriented scenarios as they emulate the natural dialogue flow between the IG and IF. A recent example of DGDS and closest to our work is Doc2Dial, a multi-domain DGDS dataset for goal-oriented dialogue modelled on hypothetical dialogue flows and dialogue scenes to simulate realistic interactions between a user and machine agent in information seeking settings \cite{feng-etal-2020-doc2dial}. Here, we follow a similar setup, however in our proposed task, we further allow users to ask clarification questions, the answers to which are are not necessarily grounded in the document. This consideration is vital in the development of instruction giving conversational agents as the dialogue pipeline needs to be more flexible, as discussed earlier. 

\subsection{Commonsense-enhanced Dialogue}
Commonsense reasoning is the innate understanding of our surroundings, situations and objects, which is essential for many AI applications \cite{ilievski2021dimensions}. Simulating these perceptual processes in task and goal oriented DGDS generates greater context and grounding for more human-like comprehension. An example of commonsense dialogue in a practical task-based scenario is understanding the common storage locations of objects, or the safe handling and use of objects from their common attributes i.e. a handle, knob or grip. Commonsense dialogue is highly contextual: In Question Answering in Context (QuAC) \cite{choi-etal-2018-quac}, dialogues are constructed from Wikipedia articles interpreted by a teacher. A student is given the title of the article and asks the teacher questions on the subject from prior knowledge, the teacher responds to the students' questions using the information in the document. This mode of question answering (Q\&A) development is more naturalistic and grounded than previous methods as the challenges of understanding the information is ingrained in the dialogue from the underlying context. Similarly, the Conversational Question Answering Challenge (CoQA) dataset \cite{reddy-etal-2019-coqa} is formulated on a rationale, scenario and conversation topic, and the Q\&As pairs are extracted from this data. This methodology is used in the Task2Dial dataset as it provides greater co-reference and pragmatic reasoning within the dialogue for enhanced comprehension as shown in Figure \ref{fig:B}.

In human-human IG/IF tasks, the IG may have prior knowledge of appropriate alternative methods, components and tools that can be used in a task that are not mentioned in the instructions. This information is vital if the IF has missing components or requires clarification on aspects of the task that are not clearly represented in the document. Variability is problematic to capture in DGDS alone as hypothetical scenarios in documents cannot account for all the potential issues in practice \cite{li-etal-2019-incremental}. Thus, the ability to ask questions that are not available in the document is crucial when conducting real-world tasks due to the changeable conditions, complexity of the task and availability of components. This is particularly important in cooking tasks (as well as other instruction giving tasks) as the user may not have all the ingredients stated in a recipe, but may have access to alternative items that can be used instead. This approach can also be used in other domains such as maintenance or construction tasks if the user does not have a specific tool but has access to a suitable alternative tool without knowing it. This inevitably introduces new challenges for dialogue systems as commonsense-related intends and actions needs to be introduced in the dialogue system. Task2Dial moves away from the closed knowledge base/s in DGDS into incorporating multiple sources of information to broaden the adaptability and application of DGDS. This is achieved by developing additional resources that listed alternative ingredients to those mentioned in the metadata from the original recipes as well as instructions on how to use cookery tools. Appropriate alternative ingredients were collected and verified using certified online cooking resources that provide food alternatives.

\begin{figure*}
    \centering
    \includegraphics[width=5.5in]{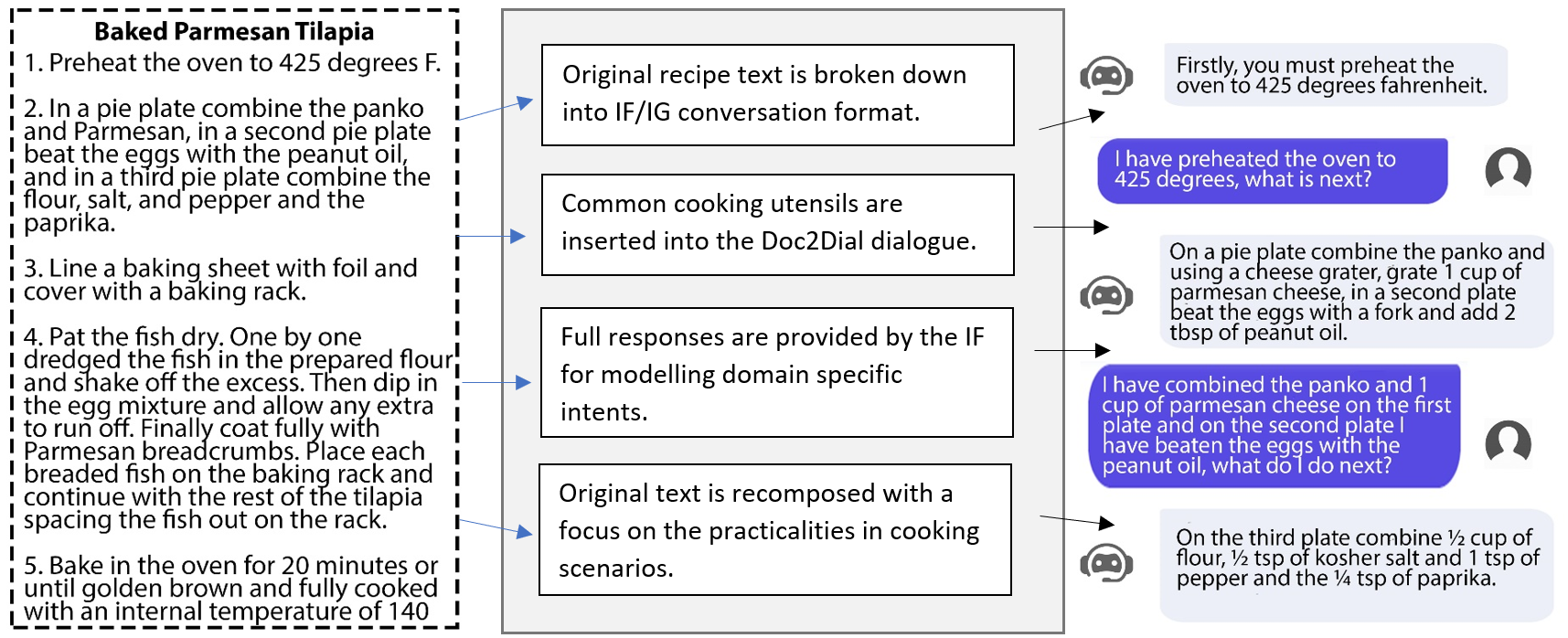}
    \caption{Original recipe text converted to Task2Dial dialogue}
    \label{fig:ds}
\end{figure*}

\section{Task2Dial} \label{sec:data-curation}

The \texttt{Task2Dial} dataset includes (1) a set of recipe documents; and (2) conversations between an IG and an IF, which are grounded in the associated recipe documents. Figure \ref{fig:ds} presents sample utterances from a dialogue along with the associated recipe. It demonstrates some important features of the dataset, such as mentioning entities not present in the recipe document; re-composition of the original text to focus on the important steps; and the break down of the recipe into manageable and appropriate steps. Following recent efforts in the field to standardise NLG research \cite{GEM}, we have made the dataset freely available\footnote{\url{www.huggingface.co/datasets/cstrathe435/Task2Dial}}.

\subsection{Data Collection Methodology}

\begin{figure}
  \includegraphics[width=\linewidth]{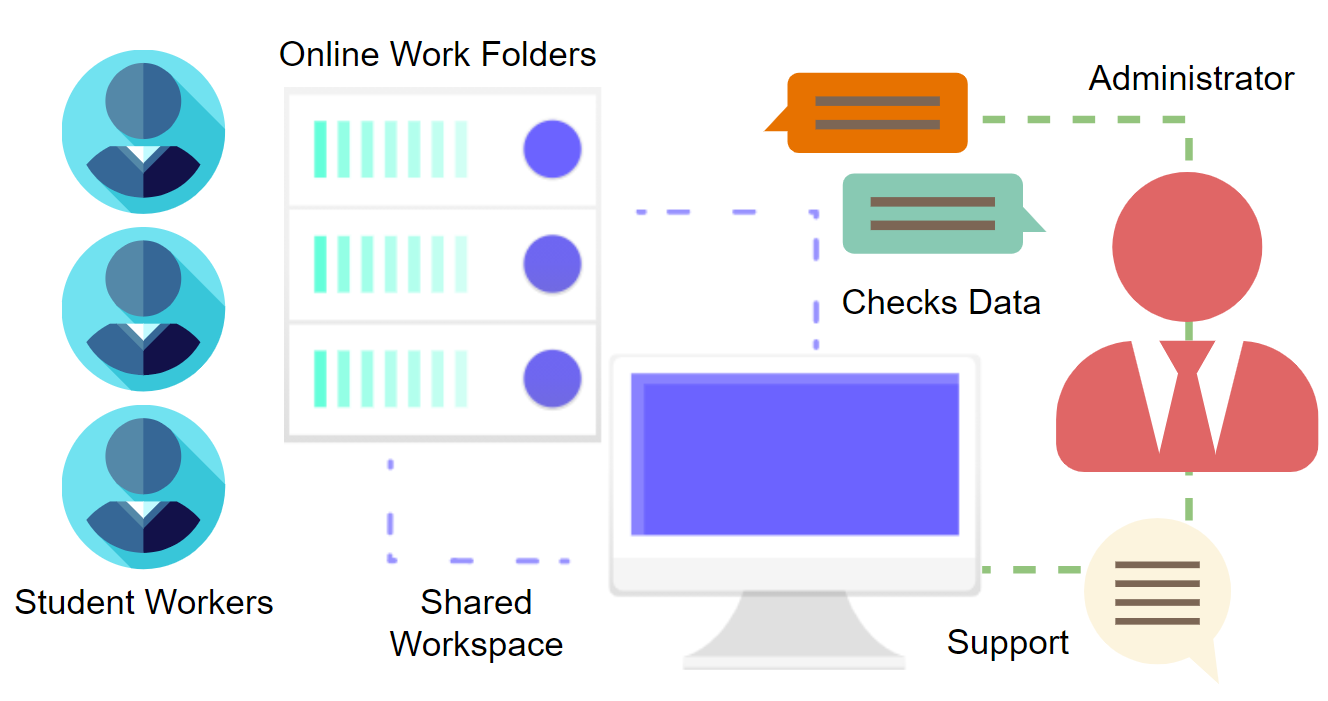}
  \caption{Overview of the Task2Dial Dataset Construction}
  \label{fig:methodology}
\end{figure}

The overall data collection methodology is shown in Figure \ref{fig:methodology} and is described in detail below. 

\paragraph{Pilot Data Collection} Prior to data collection, we performed three pilot studies. In the first, two participants assumed the roles of IG and IF respectively, where the IG had access to a recipe and provided recipe instructions to the IF (who did not have access to the recipe) over the phone, recording the session and then transcribing it. Next, we repeated the process with text-based dialogue through an online platform following a similar setup, however, the interaction was solely chat-based. The final study used \textit{self-dialogue} \cite{Byrne2019}, where one member of the team wrote entire dialogues assuming both the IF and IG roles. We found that self-dialogue results were proximal to the results of two person studies. However, time and cost was higher for producing two person dialogues, with additional time needed for transcribing and correction, thus, we opted to use self-dialogue.

\paragraph{Creation of a recipe dataset}
Three open-source and creative commons licensed cookery websites\footnote{(a) \url{www.makebetterfood.com}\\   (b) \url{www.cookeatshare.com}\\   (c) \url{www.bbcgoodfood.com}} were identified for data extraction, which permit any use or non-commercial use of data for research purposes \cite{bien-etal-2020-recipenlg,marin2019learning}. As content submission to the cooking websites was unrestricted, data appropriateness was ratified by the ratings and reviews given to each recipe by the public, highly rated recipes with positive feedback were given preference over recipes with low scores and poor reviews \cite{WANG2021138}. From this, a list of 353 recipes was compiled and divided amongst the annotators for the data collection. As mentioned earlier, annotators were asked to take on the roles of both IF and IG, rather than a multi-turn WoZ approach, to allow flexibility in the utterances. This approach allowed the annotators additional time to formulate detailed and concise responses.  
 
 \paragraph{Participants} Research assistants (RAs) from the School of Computing were employed on temporary contracts to construct and format the dataset. After an initial meeting to discuss the job role and determine suitability, the RAs were asked to complete a paid trial, this was evaluated and further advice was given on how to write dialogues and format the data to ensure high quality. After the successful completion of the trial, the RAs were permitted to continue with the remainder of the data collection. To ensure high quality of the dataset, samples of the dialogues were often reviewed and further feedback was provided. 

\paragraph{Instructions to annotators}
Each annotator was provided with a detailed list of instructions, an example dialogue and an IF/IG template (see Appendix A). The annotators were asked to read both the example dialogue and the original recipe to understand the text, context, composition, translation and annotation. The instructions included information handling and storage of data, text formatting, meta data and examples of high-quality and poor dialogues. An administrator was on hand throughout the data collection to support and guide the annotators. This approach reduced the amount of low quality dialogues associated with large crowdsourcing platforms that are often discarded post evaluation, as demonstrated in the data collection of the Doc2Dial dataset \cite{feng-etal-2020-doc2dial}.  

\paragraph{Time Scale}
The data collection was scheduled over four weeks. This was to permit additional time for the annotators to conduct work and study outside of the project. Unlike crowdsourcing methods, the annotators were given the option to work on the project flexibly in their spare time and not commit to a specific work pattern or time schedule. 

\paragraph{Ethics} 
An ethics request was submitted for review by the board of ethics at our university. No personal or other data  that may by used to identify an individual was collected in this study. 

\subsection{Task2Dial Long-form description}
\vspace{-2mm}
Unlike previous task and goal oriented DGDS, the Task2Dial corpus is unique as it is configured for practical IF/IG scenarios as demonstrated in Figure \ref{fig:ds}. Following \cite{bender-friedman-2018-data}, we provide a long-form description of the Task2Dial cooking dataset here. 

\paragraph{Curation Rationale}
Text selection was dependent on the quality of information provided in the existing recipes. Too little information and the transcription and interpretation of the text became diffused with missing or incorrect knowledge. Conversely, providing too much information in the text resulted in a lack of creativity and commonsense reasoning by the data curators. Thus, the goal of the curation was to identify text that contained all the relevant information to complete the cooking task (tools, ingredients, weights, timings, servings) but not in such detail that it subtracted from the creativity, commonsense and imagination of the annotators.


\paragraph{Language Variety} 

The recipes selected for this dataset were either written in English or translated into English prior to data collection for ease of the annotators, language understanding and future training for language models. This made the dataset accessible to all contributors involved in the curation, support and administration framework.

\paragraph{Speaker Demographics}

The recipes are composed by people of different race / ethnicity, nationalities, socioeconomic status, abilities, age, gender and language with significant variation in pronunciations, structure, language and grammar. This provided the annotators with unique linguistic content for each recipe to interpret the data and configure the text into an IF/IG format. To help preserve sociolinguistic patterns in speech, the data curators retained the underlying language when paraphrasing, to intercede social and regional dialects with their own interpretation of the data to enhance lexical richness \cite{Zampieri:2020}.

\paragraph{Annotator(s) Demographics}
Undergraduate research assistants were recruited through email. The participants were paid an hourly rate based on a university pay scale which is above the living wage and corresponds to the real living wage, following ethical guidelines for responsible innovation \cite{Silberman-responsible-2018}. 
The annotation team was composed of two males and one female data curators, under the age of 25 of mixed ethnicity's with experience in AI and computing. This minimised the gender bias that is frequently observed in crowd sourcing platforms \cite{Goodman:2012}.

\begin{figure*}
  \centering
  \begin{subfigure}[b]{0.48\linewidth}
    \includegraphics[width=\linewidth]{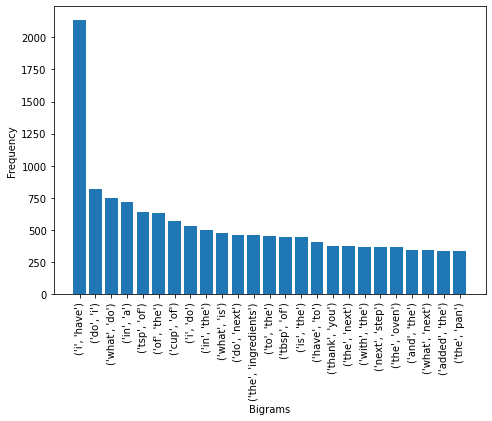}
  \end{subfigure}
  \begin{subfigure}[b]{0.48\linewidth}
    \includegraphics[width=\linewidth]{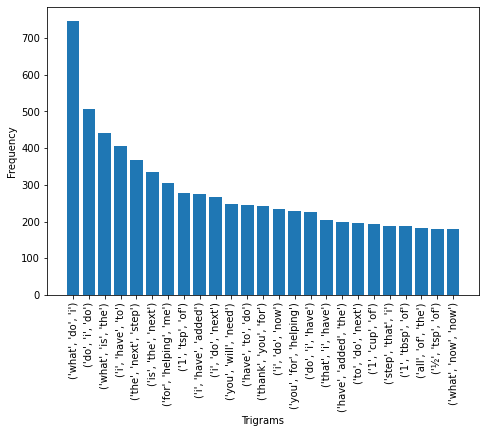}
  \end{subfigure}
  \caption{Distribution of the top 25 most frequent bigrams and trigrams in our dataset (left: most frequent bigrams, right: most frequent trigrams).}
  \label{fig:frequencies}
\end{figure*}

\begin{table*}
    \centering
        \begin{tabular}{|lccccc|}
        \hline
        Dataset   & \#docs & \#Turns & \#Tkns/Turn & TTR & MSTTR \\
        \hline \hline
        \textsc{Task2Dial} & 353 & \textbf{18.15}  &  \textbf{19.79} & \textbf{0.025} & 0.84\\
        \textsc{Doc2Dial} &  487 & 12.94  &   12 & 0.011 & 0.86 \\
        \hline
        \end{tabular}
    \caption{Size and Lexical Richness of the dataset.}
    \label{tab:statistics}
\end{table*}

\paragraph{Speech Situation} 

The annotators were given equal workloads, although workloads were adjusted accordingly over time per annotator availability to maximise data collection. The linguistic modality of the dialogue is semi-structured, synchronous interactions as existing recipes were used to paraphrase the instructions for the IG. Following this, the IF responses where created spontaneously following the logical path of the recipe in the context of the task. The intended audience for the Task2Dial dataset is broad, catering for people of different ages and abilities. Thus, the dataset is written in plain English with no jargon or unnecessary commentary to maximise accessibility. 

\paragraph{Text Characteristics} 
The structural characteristics of the Task2Dial dataset is influenced by real-world cooking scenarios that provide genre, texture and structure to the dialogues. This provides two important classifications, utterances and intents that are universal for all task-based datasets and domain specific text that is only relevant for certain tasks. This data is used when training language models as non-domain specific sample utterances such as 'I have completed this step' can be used to speed up the development of future task-based DGDS.      
\paragraph{Recording Quality}
As mentioned previously, the dialogues in Task2Dial are text-based. 

\section{Dataset Analysis} \label{sec:analysis}
This section presents overall statistics of the Task2Dial dataset. We compare our dataset to the Doc2Dial dataset, although the latter focuses on a different domain. Employing research assistants to collect and annotate data rather than using crowdsourcing platforms meant that no dialogues were discounted from the dataset. However, a pre-evaluation check was performed on the dataset before statistical analysis to reduce spelling and grammatical errors that may affect the results of the lexical analysis.
\paragraph{Size} Table \ref{tab:statistics} summarises the main descriptive statistics of Task2Dial and Doc2Dial. The dialogues in Task2Dial contain a significantly higher number of turns than Doc2Dial dialogues (18.15 as opposed to 12.94). In addition,  Task2Dial utterances are significantly longer than in Doc2Dial, containing on average more than 7 tokens. 

\paragraph{Lexical Richness \& Variation} We further report on the lexical richness and variation \cite{VanGijselSofie2005Avcl}, following \citet{novikova-etal-2017-e2e} and \citet{perez-beltrachini-gardent-2017-analysing}. We compute both Type-token ratio (TTR), i.e. the ratio of the number of word types to the number of words in a text, and the Mean segmental TTR (MSTTR), which is computed by dividing the corpus into successive segments of a given length and then calculating the average TTR of all segments to account for the fact the compared datasets are not of equal size\footnote{TTR and MSTTR have been computed using \url{https://github.com/LSYS/LexicalRichness}.}. All results are shown in Table \ref{tab:statistics}. We further investigate the distribution of the top-25 most frequent bigrams and trigrams in our dataset as seen in Figure \ref{fig:frequencies}. The majority of both trigrams (75\%) and bigrams (59\%) is only used once in the dataset, which creates a challenge to efficiently train on this data. For comparison, in Doc2Dial's 54\% of bigrams and 70\% of trigrams are used only once. Infrequent words and phrases pose a challenge for the development of data-driven dialogue systems as handling out-of-vocabulary words is a bottleneck. 

\section{Related Work} \label{sec:related-work}

This research considers the development of a DGDS for instruction-giving task-based dialogue. The work is inspired by previous research in DGDS:  Doc2Dial \cite{Ma-docdial-2020} focuses on information seeking scenarios where the interaction between an assisting agent and a user is modelled as a a sequence of dialogue scenes. To enable document-grounded dialogue, each dialogue turn consists of a dialogue scene (dialogue act, a role such as user or agent and a piece of grounding content from a document). The sequence of dialogue scenes constitute the dialogue flow. DoQA \cite{campos2020doqa} contains domain specific Q\&A dialogues in three domains including cooking, where users can ask for recommendations/instructions regarding a specific task, although the task does not involve providing steps for completing a task as well. Finally, Task2Dial has drawn inspiration from crowdsourced datasets such as MultiWoz \cite{budzianowski-etal-2018-multiwoz}, Taskmaster-1 \cite{Byrne2019} and ABCD \cite{chen-etal-2021-action} which demonstrate how DGDS can be configured in end-to-end pipelines for task-driven dialogue in virtual applications such as online booking systems. Commonsense enhanced dialogue datasets such as QuAC \cite{choi-etal-2018-quac} and CoQA \cite{reddy-etal-2019-coqa} provided key information on infusing commonsense knowledge in dialogue and commonsense actions to instil greater human-like comprehension for artificial agents to operate more effectively in the real-world.  




\section{Discussion \& Conclusions} \label{sec:conclusions}

In this paper, we introduce the Task2Dial dataset of task-based document-grounded conversations with everyday speech characteristics, between an IG and IF during a cooking task. We further extend previous work in DGDS in order to emulate the unpredictability of human-human conversations in instruction giving that do not necessarily follow a tight schema of sequential instruction giving. Instead, other discourse and dialogue phenomena might take place such as clarification questions. We further considered the aforementioned challenges of modelling dialogue for instruction-giving tasks with a focus on state-tracking, task planning, and commonsense reasoning and proposed a new task and associated dataset. 

Our proposed task aims to motivate research for modern dialogue systems that address the following challenges. Firstly, modern dialogue systems should be flexible and allow for "off-script" scenarios in order to emulate real-world phenomena, such as the ones present in human-human communication. This will require new ways of encoding user indents and new approaches to dialogue management in general. Secondly, as dialogue systems find different domain applications, the complexity of the dialogues might increase as well as the reliance of domain knowledge that can be encoded in structured or unstructured ways, such as documents, databases etc. Many applications, might require access to different domain knowledge sources in a course of a dialogue. Finally, as we design more complex dialogue systems, commonsense will play an essential part, with models required to perform reasoning with background commonsense knowledge, and generalise to tackle unseen concepts, similarly to \cite{lin-etal-2020-commongen}. In the future, we aim to benchmark and evaluate a dialogue system based on the Task2Dial dataset and the Chefbot \cite{strathearn-gkatzia-2021-chefbot}, and extend this approach to a human-robot interaction (HRI) scenario. 

\section*{Acknowledgements}
The research is supported under the EPSRC projects CiViL (EP/T014598/1) and NLG for low-resource domains (EP/T024917/1).

\bibliography{custom}

\begin{thebibliography}{38}
\expandafter\ifx\csname natexlab\endcsname\relax\def\natexlab#1{#1}\fi

\bibitem[{Bender and Friedman(2018)}]{bender-friedman-2018-data}
Emily~M. Bender and Batya Friedman. 2018.
\newblock \href {https://doi.org/10.1162/tacl_a_00041} {Data statements for
  natural language processing: Toward mitigating system bias and enabling
  better science}.
\newblock \emph{Transactions of the Association for Computational Linguistics},
  6:587--604.

\bibitem[{Bie{\'n} et~al.(2020)Bie{\'n}, Gilski, Maciejewska, Taisner,
  Wisniewski, and Lawrynowicz}]{bien-etal-2020-recipenlg}
Micha{\l} Bie{\'n}, Micha{\l} Gilski, Martyna Maciejewska, Wojciech Taisner,
  Dawid Wisniewski, and Agnieszka Lawrynowicz. 2020.
\newblock \href {https://aclanthology.org/2020.inlg-1.4} {{R}ecipe{NLG}: A
  cooking recipes dataset for semi-structured text generation}.
\newblock In \emph{Proceedings of the 13th International Conference on Natural
  Language Generation}, pages 22--28, Dublin, Ireland. Association for
  Computational Linguistics.

\bibitem[{Budzianowski et~al.(2018)Budzianowski, Wen, Tseng, Casanueva, Ultes,
  Ramadan, and Ga{\v{s}}i{\'c}}]{budzianowski-etal-2018-multiwoz}
Pawe{\l} Budzianowski, Tsung-Hsien Wen, Bo-Hsiang Tseng, I{\~n}igo Casanueva,
  Stefan Ultes, Osman Ramadan, and Milica Ga{\v{s}}i{\'c}. 2018.
\newblock \href {https://doi.org/10.18653/v1/D18-1547} {{M}ulti{WOZ} - a
  large-scale multi-domain {W}izard-of-{O}z dataset for task-oriented dialogue
  modelling}.
\newblock In \emph{Proceedings of the 2018 Conference on Empirical Methods in
  Natural Language Processing}, pages 5016--5026, Brussels, Belgium.
  Association for Computational Linguistics.

\bibitem[{Byrne et~al.(2019)Byrne, Krishnamoorthi, Sankar, Neelakantan,
  Duckworth, Yavuz, Goodrich, Dubey, Cedilnik, and Kim}]{Byrne2019}
Bill Byrne, Karthik Krishnamoorthi, Chinnadhurai Sankar, Arvind Neelakantan,
  Daniel Duckworth, Semih Yavuz, Ben Goodrich, Amit Dubey, Andy Cedilnik, and
  Kyu{-}Young Kim. 2019.
\newblock \href {http://arxiv.org/abs/1909.05358} {Taskmaster-1: Toward a
  realistic and diverse dialog dataset}.
\newblock \emph{CoRR}, abs/1909.05358.

\bibitem[{Campos et~al.(2020)Campos, Otegi, Soroa, Deriu, Cieliebak, and
  Agirre}]{campos2020doqa}
Jon~Ander Campos, Arantxa Otegi, Aitor Soroa, Jan Deriu, Mark Cieliebak, and
  Eneko Agirre. 2020.
\newblock \href {http://arxiv.org/abs/2005.01328} {Doqa -- accessing
  domain-specific faqs via conversational qa}.

\bibitem[{Chen et~al.(2021)Chen, Chen, Yang, Lin, and
  Yu}]{chen-etal-2021-action}
Derek Chen, Howard Chen, Yi~Yang, Alexander Lin, and Zhou Yu. 2021.
\newblock \href {https://www.aclweb.org/anthology/2021.naacl-main.239}
  {Action-based conversations dataset: A corpus for building more in-depth
  task-oriented dialogue systems}.
\newblock In \emph{Proceedings of the 2021 Conference of the North American
  Chapter of the Association for Computational Linguistics: Human Language
  Technologies}, pages 3002--3017, Online. Association for Computational
  Linguistics.

\bibitem[{Chen et~al.(2017)Chen, Liu, Yin, and Tang}]{Chen:2017}
Hongshen Chen, Xiaorui Liu, Dawei Yin, and Jiliang Tang. 2017.
\newblock \href {https://doi.org/10.1145/3166054.3166058} {A survey on dialogue
  systems: Recent advances and new frontiers}.
\newblock \emph{SIGKDD Explor. Newsl.}, 19(2):25–35.

\bibitem[{Choi et~al.(2018)Choi, He, Iyyer, Yatskar, Yih, Choi, Liang, and
  Zettlemoyer}]{choi-etal-2018-quac}
Eunsol Choi, He~He, Mohit Iyyer, Mark Yatskar, Wen-tau Yih, Yejin Choi, Percy
  Liang, and Luke Zettlemoyer. 2018.
\newblock {Q}u{AC}: Question answering in context.
\newblock In \emph{Proceedings of the 2018 Conference on Empirical Methods in
  Natural Language Processing}.

\bibitem[{Clinciu et~al.(2021)Clinciu, Gkatzia, and
  Mahamood}]{clinciu-etal-2021-commonsense}
Miruna-Adriana Clinciu, Dimitra Gkatzia, and Saad Mahamood. 2021.
\newblock \href {https://aclanthology.org/2021.humeval-1.1} {It{'}s
  commonsense, isn{'}t it? demystifying human evaluations in
  commonsense-enhanced {NLG} systems}.
\newblock In \emph{Proceedings of the Workshop on Human Evaluation of NLP
  Systems (HumEval)}, pages 1--12, Online. Association for Computational
  Linguistics.

\bibitem[{Feng et~al.(2020)Feng, Wan, Gunasekara, Patel, Joshi, and
  Lastras}]{feng-etal-2020-doc2dial}
Song Feng, Hui Wan, Chulaka Gunasekara, Siva Patel, Sachindra Joshi, and Luis
  Lastras. 2020.
\newblock \href {https://doi.org/10.18653/v1/2020.emnlp-main.652} {doc2dial: A
  goal-oriented document-grounded dialogue dataset}.
\newblock In \emph{Proceedings of the 2020 Conference on Empirical Methods in
  Natural Language Processing (EMNLP)}, pages 8118--8128, Online. Association
  for Computational Linguistics.

\bibitem[{Gargett et~al.(2010)Gargett, Garoufi, Koller, and
  Striegnitz}]{gargett-etal-2010-give}
Andrew Gargett, Konstantina Garoufi, Alexander Koller, and Kristina Striegnitz.
  2010.
\newblock \href
  {http://www.lrec-conf.org/proceedings/lrec2010/pdf/532_Paper.pdf} {The
  {GIVE}-2 corpus of giving instructions in virtual environments}.
\newblock In \emph{Proceedings of the Seventh International Conference on
  Language Resources and Evaluation ({LREC}'10)}, Valletta, Malta. European
  Language Resources Association (ELRA).

\bibitem[{Gehrmann et~al.(2021)Gehrmann, Adewumi, Aggarwal, Ammanamanchi,
  Anuoluwapo, Bosselut, Chandu, Clinciu, Das, Dhole, Du, Durmus, Dusek, Emezue,
  Gangal, Garbacea, Hashimoto, McMillan{-}Major, Mille, van Miltenburg, Nadeem,
  Narayan, Nikolaev, and Niyongabo}]{GEM}
Sebastian Gehrmann, Tosin~P. Adewumi, Karmanya Aggarwal, Pawan~Sasanka
  Ammanamanchi, Aremu Anuoluwapo, Antoine Bosselut, Khyathi~Raghavi Chandu,
  Miruna{-}Adriana Clinciu, Dipanjan Das, Kaustubh~D. Dhole, Wanyu Du, Esin
  Durmus, Ondrej Dusek, Chris Emezue, Varun Gangal, Cristina Garbacea,
  Tatsunori Hashimoto, Angelina McMillan{-}Major, Simon Mille, Emiel van
  Miltenburg, Moin Nadeem, Shashi Narayan, Vitaly Nikolaev, and Rubungo~Andre
  Niyongabo. 2021.
\newblock \href {http://arxiv.org/abs/2102.01672} {The {GEM} benchmark: Natural
  language generation, its evaluation and metrics}.
\newblock \emph{CoRR}, abs/2102.01672.

\bibitem[{Gkatzia and Belvedere(2021)}]{Gkatzia-hri-2021}
Dimitra Gkatzia and Francesco Belvedere. 2021.
\newblock "what's this?" comparing active learning strategies for concept
  acquisition in hri.
\newblock In \emph{Companion of the 2021 ACM/IEEE International Conference on
  Human-Robot Interaction}, HRI '21 Companion, page 205–209.

\bibitem[{Goodman et~al.(2012)Goodman, Cryder, and Cheema}]{Goodman:2012}
Joseph~K. Goodman, Cynthia Cryder, and Amar Cheema. 2012.
\newblock Data collection in a flat world: Strengths and weaknesses of
  mechanical turk samples.
\newblock \emph{Journal of Behavioral Decision Making, Forthcoming}.

\bibitem[{Ham et~al.(2020)Ham, Lee, Jang, and Kim}]{ham-etal-2020-end}
Donghoon Ham, Jeong-Gwan Lee, Youngsoo Jang, and Kee-Eung Kim. 2020.
\newblock End-to-end neural pipeline for goal-oriented dialogue systems using
  {GPT}-2.
\newblock In \emph{Proceedings of the 58th Annual Meeting of the Association
  for Computational Linguistics}.

\bibitem[{Hosseini-Asl et~al.(2020)Hosseini-Asl, McCann, Wu, Yavuz, and
  Socher}]{NEURIPS2020_e9462095}
Ehsan Hosseini-Asl, Bryan McCann, Chien-Sheng Wu, Semih Yavuz, and Richard
  Socher. 2020.
\newblock \href
  {https://proceedings.neurips.cc/paper/2020/file/e946209592563be0f01c844ab2170f0c-Paper.pdf}
  {A simple language model for task-oriented dialogue}.
\newblock In \emph{Advances in Neural Information Processing Systems},
  volume~33, pages 20179--20191. Curran Associates, Inc.

\bibitem[{Hu et~al.(2016)Hu, Dick, Chang, Bowden, Neff, Fox~Tree, and
  Walker}]{hu-etal-2016-corpus}
Zhichao Hu, Michelle Dick, Chung-Ning Chang, Kevin Bowden, Michael Neff, Jean
  Fox~Tree, and Marilyn Walker. 2016.
\newblock \href {https://aclanthology.org/L16-1550} {A corpus of
  gesture-annotated dialogues for monologue-to-dialogue generation from
  personal narratives}.
\newblock In \emph{Proceedings of the Tenth International Conference on
  Language Resources and Evaluation ({LREC}'16)}, pages 3447--3454,
  Portoro{\v{z}}, Slovenia. European Language Resources Association (ELRA).

\bibitem[{Ilievski et~al.(2021)Ilievski, Oltramari, Ma, Zhang, McGuinness, and
  Szekely}]{ilievski2021dimensions}
Filip Ilievski, Alessandro Oltramari, Kaixin Ma, Bin Zhang, Deborah~L.
  McGuinness, and Pedro Szekely. 2021.
\newblock \href {http://arxiv.org/abs/2101.04640} {Dimensions of commonsense
  knowledge}.

\bibitem[{Ilievski et~al.(2018)Ilievski, Musat, Hossmann, and
  Baeriswyl}]{Ilievski:2018}
Vladimir Ilievski, Claudiu Musat, Andreea Hossmann, and Michael Baeriswyl.
  2018.
\newblock Goal-oriented chatbot dialog management bootstrapping with transfer
  learning.
\newblock In \emph{Proceedings of the 27th International Joint Conference on
  Artificial Intelligence}, IJCAI'18, page 4115–4121. AAAI Press.

\bibitem[{Li et~al.(2019)Li, Niu, Meng, Feng, Li, and
  Zhou}]{li-etal-2019-incremental}
Zekang Li, Cheng Niu, Fandong Meng, Yang Feng, Qian Li, and Jie Zhou. 2019.
\newblock \href {https://doi.org/10.18653/v1/P19-1002} {Incremental transformer
  with deliberation decoder for document grounded conversations}.
\newblock In \emph{Proceedings of the 57th Annual Meeting of the Association
  for Computational Linguistics}, pages 12--21, Florence, Italy. Association
  for Computational Linguistics.

\bibitem[{Lin et~al.(2020)Lin, Zhou, Shen, Zhou, Bhagavatula, Choi, and
  Ren}]{lin-etal-2020-commongen}
Bill~Yuchen Lin, Wangchunshu Zhou, Ming Shen, Pei Zhou, Chandra Bhagavatula,
  Yejin Choi, and Xiang Ren. 2020.
\newblock \href {https://doi.org/10.18653/v1/2020.findings-emnlp.165}
  {{C}ommon{G}en: A constrained text generation challenge for generative
  commonsense reasoning}.
\newblock In \emph{Findings of the Association for Computational Linguistics:
  EMNLP 2020}, pages 1823--1840, Online. Association for Computational
  Linguistics.

\bibitem[{Ma et~al.(2020)Ma, Zhang, Li, and Liu}]{Ma-docdial-2020}
Longxuan Ma, Wei{-}Nan Zhang, Mingda Li, and Ting Liu. 2020.
\newblock \href {http://arxiv.org/abs/2004.13818} {A survey of document
  grounded dialogue systems {(DGDS)}}.
\newblock \emph{CoRR}, abs/2004.13818.

\bibitem[{Marin et~al.(2019)Marin, Biswas, Ofli, Hynes, Salvador, Aytar, Weber,
  and Torralba}]{marin2019learning}
Javier Marin, Aritro Biswas, Ferda Ofli, Nicholas Hynes, Amaia Salvador, Yusuf
  Aytar, Ingmar Weber, and Antonio Torralba. 2019.
\newblock Recipe1m+: A dataset for learning cross-modal embeddings for cooking
  recipes and food images.
\newblock \emph{{IEEE} Trans. Pattern Anal. Mach. Intell.}

\bibitem[{Novikova et~al.(2017)Novikova, Du{\v{s}}ek, and
  Rieser}]{novikova-etal-2017-e2e}
Jekaterina Novikova, Ond{\v{r}}ej Du{\v{s}}ek, and Verena Rieser. 2017.
\newblock \href {https://doi.org/10.18653/v1/W17-5525} {The {E}2{E} dataset:
  New challenges for end-to-end generation}.
\newblock In \emph{Proceedings of the 18th Annual {SIG}dial Meeting on
  Discourse and Dialogue}, pages 201--206, Saarbr{\"u}cken, Germany.
  Association for Computational Linguistics.

\bibitem[{Panagiaris et~al.(2021)Panagiaris, Hart, and
  Gkatzia}]{PANAGIARIS2021101184}
Nikolaos Panagiaris, Emma Hart, and Dimitra Gkatzia. 2021.
\newblock \href {https://doi.org/https://doi.org/10.1016/j.csl.2020.101184}
  {Generating unambiguous and diverse referring expressions}.
\newblock \emph{Computer Speech \& Language}, 68:101184.

\bibitem[{Perez-Beltrachini and
  Gardent(2017)}]{perez-beltrachini-gardent-2017-analysing}
Laura Perez-Beltrachini and Claire Gardent. 2017.
\newblock \href {https://doi.org/10.18653/v1/W17-3537} {Analysing data-to-text
  generation benchmarks}.
\newblock In \emph{Proceedings of the 10th International Conference on Natural
  Language Generation}, pages 238--242, Santiago de Compostela, Spain.
  Association for Computational Linguistics.

\bibitem[{Reddy et~al.(2019)Reddy, Chen, and Manning}]{reddy-etal-2019-coqa}
Siva Reddy, Danqi Chen, and Christopher~D. Manning. 2019.
\newblock \href {https://doi.org/10.1162/tacl_a_00266} {{C}o{QA}: A
  conversational question answering challenge}.
\newblock \emph{Transactions of the Association for Computational Linguistics},
  7:249--266.

\bibitem[{Shah et~al.(2018)Shah, Hakkani-T{\"u}r, Liu, and
  T{\"u}r}]{shah-etal-2018-bootstrapping}
Pararth Shah, Dilek Hakkani-T{\"u}r, Bing Liu, and Gokhan T{\"u}r. 2018.
\newblock \href {https://doi.org/10.18653/v1/N18-3006} {Bootstrapping a neural
  conversational agent with dialogue self-play, crowdsourcing and on-line
  reinforcement learning}.
\newblock In \emph{Proceedings of the 2018 Conference of the North {A}merican
  Chapter of the Association for Computational Linguistics: Human Language
  Technologies, Volume 3 (Industry Papers)}, pages 41--51, New Orleans -
  Louisiana. Association for Computational Linguistics.

\bibitem[{Shum et~al.(2018)Shum, He, and Li}]{shum2018eliza}
Heung-Yeung Shum, Xiaodong He, and Di~Li. 2018.
\newblock \href {http://arxiv.org/abs/1801.01957} {From eliza to xiaoice:
  Challenges and opportunities with social chatbots}.

\bibitem[{Silberman et~al.(2018)Silberman, Tomlinson, LaPlante, Ross, Irani,
  and Zaldivar}]{Silberman-responsible-2018}
M.~S. Silberman, B.~Tomlinson, R.~LaPlante, J.~Ross, L.~Irani, and A.~Zaldivar.
  2018.
\newblock \href {https://doi.org/10.1145/3180492} {Responsible research with
  crowds: Pay crowdworkers at least minimum wage}.
\newblock \emph{Commun. ACM}, 61(3):39–41.

\bibitem[{Stoyanchev and Piwek(2010)}]{stoyanchev-piwek-2010-constructing}
Svetlana Stoyanchev and Paul Piwek. 2010.
\newblock Constructing the {CODA} corpus: A parallel corpus of monologues and
  expository dialogues.
\newblock In \emph{Proceedings of the Seventh International Conference on
  Language Resources and Evaluation ({LREC}'10)}, Valletta, Malta. European
  Language Resources Association (ELRA).

\bibitem[{Strathearn and Gkatzia(2021)}]{strathearn-gkatzia-2021-chefbot}
Carl Strathearn and Dimitra Gkatzia. 2021.
\newblock \href {https://aclanthology.org/2021.inlg-1.5} {Chefbot: A novel
  framework for the generation of commonsense-enhanced responses for task-based
  dialogue systems}.
\newblock In \emph{Proceedings of the 14th International Conference on Natural
  Language Generation}, pages 46--47, Aberdeen, Scotland, UK. Association for
  Computational Linguistics.

\bibitem[{Van~Gijsel et~al.(2005)Van~Gijsel, Speelman, and
  Geeraerts}]{VanGijselSofie2005Avcl}
Sofie Van~Gijsel, Dirk Speelman, and Dirk Geeraerts. 2005.
\newblock A variationist, corpus linguistic analysis of lexical richness.
\newblock In \emph{Proceedings from the Corpus Linguistics Conference Series},
  volume~1, pages 1--16.

\bibitem[{Wang and Kim(2021)}]{WANG2021138}
Yiqi Wang and Jewoo Kim. 2021.
\newblock \href {https://doi.org/https://doi.org/10.1016/j.jhtm.2021.05.016}
  {Interconnectedness between online review valence, brand, and restaurant
  performance}.
\newblock \emph{Journal of Hospitality and Tourism Management}, 48:138--145.

\bibitem[{Zamanirad et~al.(2020)Zamanirad, Benatallah, Rodriguez,
  Yaghoubzadehfard, Bouguelia, and Brabra}]{Zamanirad2020}
Shayan Zamanirad, Boualem Benatallah, Carlos Rodriguez, Mohammadali
  Yaghoubzadehfard, Sara Bouguelia, and Hayet Brabra. 2020.
\newblock State machine based human-bot conversation model and services.
\newblock In \emph{Advanced Information Systems Engineering}, pages 199--214,
  Cham. Springer International Publishing.

\bibitem[{Zampieri et~al.(2020)Zampieri, Nakov, and Scherrer}]{Zampieri:2020}
Marcos Zampieri, Preslav Nakov, and Yves Scherrer. 2020.
\newblock \href {https://doi.org/10.1017/S1351324920000492} {Natural language
  processing for similar languages, varieties, and dialects: A survey}.
\newblock \emph{Natural Language Engineering}, 26(6):595–612.

\bibitem[{Zhang et~al.(2020)Zhang, Takanobu, Huang, and
  Zhu}]{Zhang-recent-2020}
Zheng Zhang, Ryuichi Takanobu, Minlie Huang, and Xiaoyan Zhu. 2020.
\newblock \href {http://arxiv.org/abs/2003.07490} {Recent advances and
  challenges in task-oriented dialog system}.
\newblock \emph{CoRR}, abs/2003.07490.

\bibitem[{Zhou et~al.(2018)Zhou, Prabhumoye, and
  Black}]{zhou-etal-2018-dataset}
Kangyan Zhou, Shrimai Prabhumoye, and Alan~W Black. 2018.
\newblock \href {https://doi.org/10.18653/v1/D18-1076} {A dataset for document
  grounded conversations}.
\newblock In \emph{Proceedings of the 2018 Conference on Empirical Methods in
  Natural Language Processing}, pages 708--713, Brussels, Belgium. Association
  for Computational Linguistics.

\end{thebibliography}
\bibliographystyle{acl_natbib}

\end{document}